\documentclass[letterpaper,10pt,conference]{ieeeconf}

\IEEEoverridecommandlockouts
\usepackage{amsmath,amssymb,amsfonts,bm,mathtools}
\usepackage{graphicx}
\usepackage{booktabs}
\usepackage{subfig}
\usepackage{cite}
\usepackage{xcolor}
\usepackage{hyperref}
\usepackage{array}
\usepackage{algorithm}
\usepackage{algpseudocode}
\usepackage{balance}

\hypersetup{
  colorlinks=true,
  linkcolor=blue,
  citecolor=orange,
  urlcolor=black
}

\newcommand{\normtwo}[1]{\left\lVert #1 \right\rVert_2}
\newcommand{\figinc}[2]{%
  \IfFileExists{#1}{\includegraphics[width=#2]{#1}}%
  {\fbox{\parbox[c][3.1cm][c]{#2}{\centering Missing figure\\\texttt{\detokenize{#1}}}}}%
}

\title{\LARGE \bf
Real-Time Safety Evaluation of Human Arm Operations Using a Wrist-Mounted IMU with PSM System
}

\author{
Musab Zubair Inamdar, Alhusain Al Hadrami, and Seyed Amir Tafrishi%
\thanks{Musab Zubair Inamdar and Seyed Amir Tafrishi are with the Geometric Mechanics and Mechatronics in Robotics (gm$^2$R) Lab, School of Engineering, Cardiff University, Cardiff CF24 3AA, United Kingdom. Email: musabinamdar@gmail.com, AlhadramiA@cardiff.ac.uk and tafrishisa@cardiff.ac.uk}
}
\renewcommand{\baselinestretch}{.9}
\begin{document}
\maketitle
\thispagestyle{empty}
\pagestyle{empty}

\begin{abstract}
This paper presents a wrist-specific adaptation of the Predictive Safety Model (PSM) for real-time evaluation of human arm operations in collaborative manufacturing. A wrist-mounted inertial measurement unit (IMU) is used to measure wrist orientation and angular velocity during three representative manual tasks: fastening with a hand tool, visual inspection, and pick-and-place. The proposed method combines a spring-damper-mass predictive model, which provides a task-consistent safe reference motion, with an impedance-inspired safety index that fuses orientation and angular-velocity discrepancies for online assessment. In addition, a spectral error indicator is retained as an offline comparative baseline. Experimental validation over 18 task trials shows a clear separation between controlled motions and deliberately irregular or hazardous-like motions across all three tasks. The results further indicate that velocity-weighted impedance evaluation is more stable than position-weighted evaluation for wrist-mounted IMU sensing. The present study is intended as a proof-of-concept validation of wearable safety assessment rather than a population-scale study, but it demonstrates that a lightweight wrist sensor and an interpretable predictive model can provide useful real-time safety information for human-robot collaborative workcells.
\end{abstract}

\section{Introduction}


Collaborative robotics has become a major enabling technology in modern manufacturing by allowing humans and robots to work in shared spaces with improved flexibility, productivity, and task adaptability \cite{vicentini2021collaborative,ravankar2022care,villani2018survey}. However, as collaborative systems become closer to human operators, safety cannot be treated as a purely robot-side problem. In addition to robot speed, force, and separation distance, the quality and regularity of the human motion itself become important, especially during manual operations that involve repeated wrist rotations, abrupt corrections, or fine manipulation near tools and parts \cite{lasota2017survey,robla2017working,gualtieri2021emerging}.

Industrial safety standards provide a necessary basis for safe deployment of collaborative robots, yet they are often implemented through conservative workspace-level safeguards or robot-side constraints \cite{davisonsafety,marvel2017implementing,michalos2015design}. Such strategies are important, but they do not directly quantify whether the operator's arm motion is itself controlled, irregular, or potentially hazardous during task execution. This limitation is particularly relevant in collaborative manufacturing cells, where human behaviour is highly task dependent and may change quickly due to fatigue, distraction, poor alignment, or rushed operation \cite{segura2022safety,benmessabih2024online,pupa2023dynamic}.

A large part of the literature on human-robot safety has therefore focused on motion prediction, collision avoidance, dynamic planning, and robot adaptation to human presence \cite{pedrocchi2013safe,zanchettin2015safety,unhelkar2018human,chan2020collision,kimmel2017invariance}. Wearable inertial sensing, on the other hand, has been used widely in human motion analysis and activity monitoring because of its low cost, portability, and direct access to limb kinematics \cite{prakash2018recent,benmessabih2024online}. Nevertheless, the use of a wearable IMU as a continuous wrist-level safety evaluator for collaborative manufacturing remains comparatively underexplored. In particular, there is still a need for a compact and interpretable safety metric that can be computed directly from wearable motion data without instrumenting the full workspace.

This paper builds on the Predictive Safety Model (PSM) introduced in \cite{tafrishi2022psm}, where human upper-body motion was represented by a spring-damper pendulum model. Here, the same core idea is adapted to wrist and forearm motion relevant to manufacturing operations. The contribution of this paper is therefore not a completely new human motion model, but a wrist-specific formulation and evaluation pipeline that makes the earlier PSM approach more directly applicable to human-robot collaborative tasks. In particular, we combine the predictive model with an impedance-inspired error index that merges orientation and angular-velocity discrepancies in the time domain, while retaining a spectral error analysis as an offline comparative baseline.

The main contributions of this work are as follows:
\begin{itemize}
    \item A wrist-mounted IMU-based adaptation of the spring-damper-mass Predictive Safety Model for human arm operations in collaborative manufacturing.
    \item A self-contained real-time evaluation pipeline that maps wrist motion to a probabilistic safe-reference model and computes safety from orientation and angular-velocity deviations.
    \item An impedance-inspired safety index that addresses the practical limitations of relying only on frequency-domain indicators for online assessment.
    \item Experimental validation on three representative tasks---fastening with a hand tool, visual inspection, and pick-and-place---together with a discussion of threshold selection, IMU noise sensitivity, and current limitations.
\end{itemize}

To clarify the scope of novelty, the core spring-damper Predictive Safety Model is inherited from our earlier upper-body formulation in \cite{tafrishi2022psm}. The present contribution is a wrist-level adaptation for collaborative manufacturing, including wrist-specific parameterisation, a reduced wrist-motion probability representation, and an impedance-inspired online safety index with velocity-priority evaluation. The contribution of this paper is therefore a practical wrist-specific evaluation framework built on the original PSM, rather than a completely new predictive model. The remainder of this paper is organized as follows. Section~\ref{sec:Design} presents the wrist-specific PSM formulation, sensing pipeline, reduced probability-map construction, and safety metrics. Section~\ref{sec:exp_method} describes the experimental protocol and task setup. Section~\ref{sec:motion_study} reports the results and discussion. Finally, Section~\ref{sec:conclusions} concludes the paper and outlines future work.

\section{Predictive Safety Model for Wrist Motion}
\label{sec:Design}


\subsection{System Overview and Wrist-Specific PSM Formulation}

\begin{figure}[t!]
    \centering
    \figinc{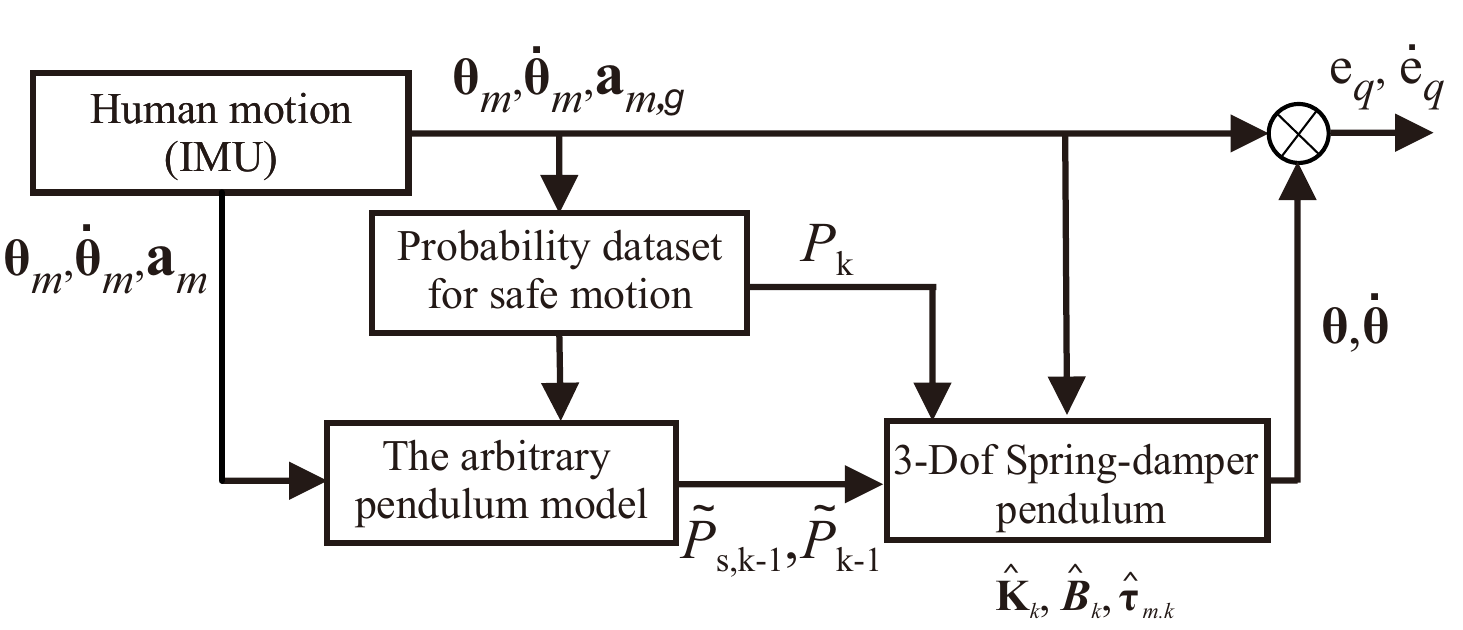}{0.5\textwidth}
    \caption{Overview of the wrist-level predictive safety evaluation pipeline. The IMU provides the measured wrist motion, the probability map and predictive model generate a task-consistent safe reference, and the final safety index is computed from the discrepancy between the measured and predicted motions.}
    \label{fig:PSMblock}
\end{figure}

The proposed system evaluates the safety of human arm operations from a wrist-mounted IMU and a wrist-specific adaptation of the Predictive Safety Model (PSM) in \cite{tafrishi2022psm}. The sensing and evaluation architecture is illustrated in Fig.~\ref{fig:PSMblock}. At each time step, the IMU provides the measured wrist orientation and angular velocity, while the PSM generates a task-consistent safe reference motion with frequency of 20 Hz. Safety is then evaluated from the discrepancy between the measured motion and the predicted safe motion.

\begin{figure}[t!]
    \centering
    \figinc{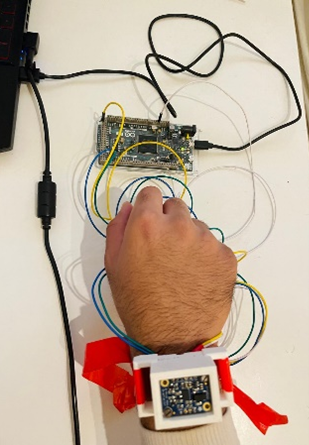}{0.17\textwidth}\hfill
    \figinc{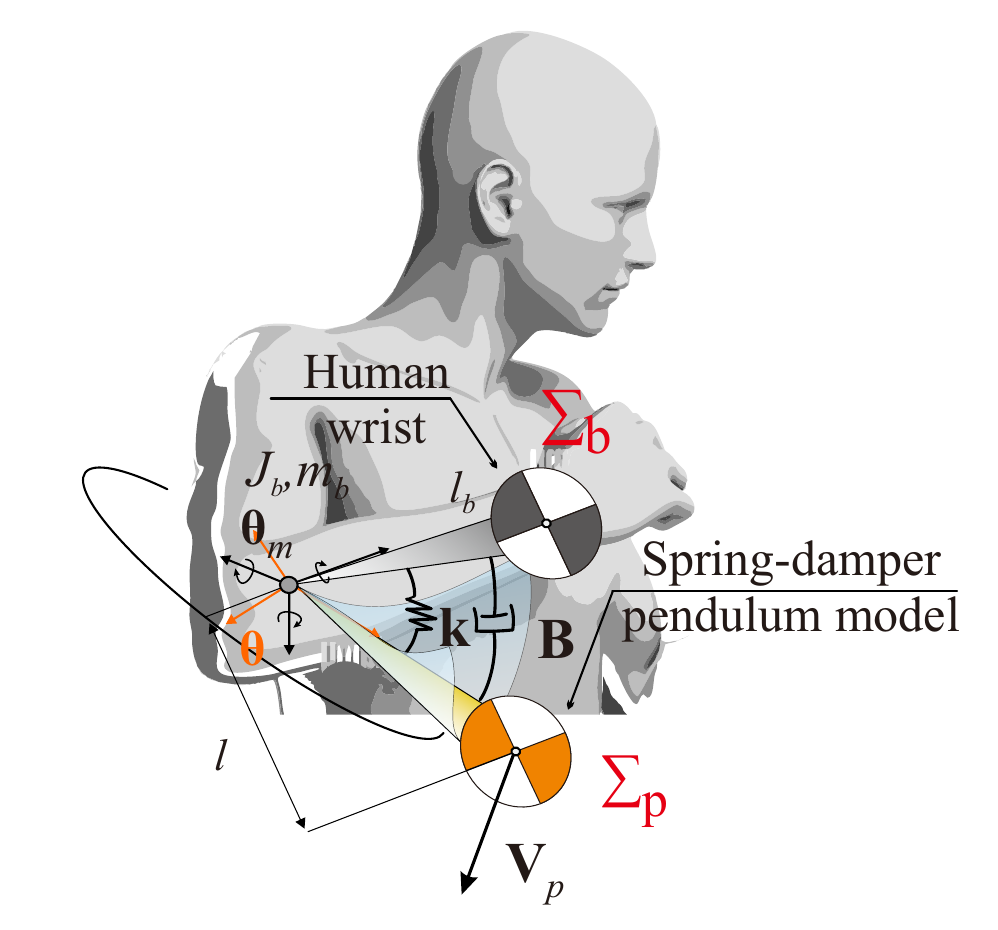}{0.30\textwidth}
    \caption{Wrist-mounted IMU prototype and the corresponding wrist-level spring-damper-mass abstraction used in the predictive safety model.}
    \label{fig:IMU_setup}
\end{figure}

Following the previous PSM framework, the wrist motion is represented by a reduced spring-damper-mass model,
\begin{equation}
\mathbf{M}(\bm{\theta})\ddot{\bm{\theta}}
+\mathbf{H}(\bm{\theta},\dot{\bm{\theta}})
+\mathbf{G}(\bm{\theta})
+\mathbf{B}\!\left(\dot{\bm{\theta}}-\dot{\bm{\theta}}_m\right)
+\mathbf{K}\!\left(\bm{\theta}-\bm{\theta}_m\right)
=
\bm{\tau}_m,
\label{eq:psm_wrist}
\end{equation}
where $\bm{\theta}=[\theta_x,\theta_y,\theta_z]^\top$ and $\dot{\bm{\theta}}$ are the wrist-state variables predicted by the PSM, and $\bm{\theta}_m=[\theta_{m,x},\theta_{m,y},\theta_{m,z}]^\top$ and $\dot{\bm{\theta}}_m$ are the measured wrist orientation and angular velocity obtained from the IMU. The matrices $\mathbf{M}$, $\mathbf{H}$, and $\mathbf{G}$ represent the inertia, nonlinear, and gravity terms, while $\mathbf{K}$ and $\mathbf{B}$ denote the effective stiffness and damping terms that connect the predictive model to the measured motion. The generalized input $\bm{\tau}_m$ represents the measured motion effect driving the predictive model.

In contrast to \cite{tafrishi2022psm}, which focused on upper-body motion, the present formulation is tuned for wrist and forearm motion relevant to manufacturing tasks. The effective physical parameters were reduced accordingly, with an arm mass of $2~\mathrm{kg}$, an effective pendulum length of $0.2~\mathrm{m}$, and a base radius of $0.25~\mathrm{m}$. The nominal stiffness and damping vectors were set to
\[
\mathbf{K}_0=[500,\,500,\,1200]^\top,\qquad
\mathbf{B}_0=[40,\,40,\,60]^\top,
\]
which provided stable and repeatable behaviour across the three tasks considered in this study. These values should be interpreted as experimentally chosen effective model parameters for the present wrist-level proof-of-concept rather than universal biomechanical constants.

\subsection{IMU Processing and Reduced Probability Map}

A BNO055 IMU was mounted on the participant's wrist, as shown in Fig.~\ref{fig:IMU_setup}. The current prototype is a wired data-acquisition setup connected to an Arduino-based interface; therefore, in this paper the device is referred to as \emph{wrist-mounted} rather than wireless. The IMU provides fused orientation estimates and angular-velocity measurements, which are used as the primary motion signals for safety evaluation. Linear acceleration was also recorded as an auxiliary signal, but it was not used directly in the final impedance-based safety index.

Before each experiment, a short neutral-pose calibration was performed to define the reference sensor frame. Let $\mathbf{R}_m(t)\in SO(3)$ denote the orientation of the wrist-mounted sensor with respect to the calibrated global frame, and let $\hat{\mathbf{u}}_s$ be the calibrated sensor axis associated with the dominant wrist direction. We define the gravity-referenced angular coordinate as
\begin{equation}
\theta_g(t)=\cos^{-1}\!\left(\mathbf{e}_3^\top \mathbf{R}_m(t)\hat{\mathbf{u}}_s\right),
\label{eq:theta_g}
\end{equation}
where $\mathbf{e}_3=[0\;0\;1]^\top$ denotes the global vertical direction. The angular-speed coordinate is taken as the Euclidean norm of the measured wrist angular velocity,
\begin{equation}
\omega(t)=\normtwo{\dot{\bm{\theta}}_m(t)}.
\label{eq:omega_def}
\end{equation}
Hence, all three rotational axes are used in the safety evaluation, while task-level plots are still shown per axis for interpretation.

The IMU was configured in AMG mode and streamed to the host through the Arduino interface. The predictive safety evaluator operated on a filtered discrete-time data stream. A low-pass filter with cutoff below the Nyquist frequency of the processed data rate was applied to suppress sensor noise without masking task-relevant wrist dynamics.


To connect the measured wrist motion with the reduced PSM representation, we use the same reduced state variables as in \cite{tafrishi2022psm}: the gravity-referenced angular coordinate $\theta_g$ and the angular-speed magnitude $\omega$. A task-specific probability map is then formed from recorded trials as
\begin{equation}
P_{ij}=P(\theta_{g,i},\omega_j),
\label{eq:prob_map}
\end{equation}
where each entry stores the empirical likelihood of observing the motion state $(\theta_g,\omega)$ during task execution. High-probability regions correspond to regular and repeatable wrist motion patterns, whereas low-probability regions typically correspond to irregular motion, abrupt corrections, or uncommon wrist configurations.

In real time, the safety evaluator operates in four steps:
\begin{enumerate}
    \item acquire the current IMU-based orientation and angular velocity,
    \item map the current motion to the reduced state $(\theta_g,\omega)$,
    \item query the local probability neighbourhood to determine the most likely safe continuation of the task motion,
    \item propagate the PSM in (\ref{eq:psm_wrist}) and obtain the predicted safe state $(\bm{\theta},\dot{\bm{\theta}})$.
\end{enumerate}
This procedure keeps the online computation lightweight because it uses a reduced probability lookup and a low-order predictive model instead of a high-dimensional optimization or full-body estimation pipeline.

\subsection{Online Safety Indicators}

For online evaluation, the discrepancy between the measured wrist motion and the PSM prediction is computed in the time domain as
\begin{equation}
e_{\theta}(k)=\normtwo{\bm{\theta}_m(k)-\bm{\theta}(k)},
e_{\omega}(k)=\normtwo{\dot{\bm{\theta}}_m(k)-\dot{\bm{\theta}}(k)},
\label{eq:time_errors}
\end{equation}
where $k$ denotes the discrete-time index. These two signals quantify, respectively, the instantaneous orientation mismatch and angular-velocity mismatch between the human motion and the predicted safe motion.

To obtain a direct online safety measure, we define an impedance-inspired normalized error metric
\begin{equation}
e_{\mathrm{imp}}(k)=
P^*\frac{e_{\theta}(k)}{\bar e_{\theta}}
+
D^*\frac{e_{\omega}(k)}{\bar e_{\omega}},
\label{eq:impedance_metric}
\end{equation}
where $\bar e_{\theta}$ and $\bar e_{\omega}$ are normalization constants computed from the calibration dataset of the corresponding task, and $P^*$ and $D^*$ are the relative weights assigned to orientation and angular-velocity errors, respectively. In the present study, normalization was performed using the maximum observed error of each task-specific calibration set so that $e_{\mathrm{imp}}$ remained dimensionless and comparable across trials.

Two weighting cases were considered:
\begin{equation}
\begin{cases}
\textnormal{position-priority:} & P^*=0.6,\; D^*=0.4,\\
\textnormal{velocity-priority:} & P^*=0.4,\; D^*=0.6.
\end{cases}
\label{eq:weight_cases}
\end{equation}
The motivation is that wrist-mounted IMUs are often more robust to short-term angular-velocity changes than to small drift in absolute orientation estimates; therefore, the velocity-priority case can be more stable in practice.

The corresponding online safety level is defined as
\begin{equation}
S(k)=
\begin{cases}
\textnormal{safe}, & e_{\mathrm{imp}}(k)<\varepsilon_s,\\
\textnormal{caution}, & \varepsilon_s \le e_{\mathrm{imp}}(k)<\varepsilon_c,\\
\textnormal{unsafe-like}, & e_{\mathrm{imp}}(k)\ge \varepsilon_c,
\end{cases}
\label{eq:safety_levels}
\end{equation}
where $\varepsilon_s$ and $\varepsilon_c$ are empirical operating thresholds chosen from controlled and deliberately irregular task executions. In the present dataset, controlled operation mostly remained below $\varepsilon_s=0.15$, while irregular wrist events consistently crossed $\varepsilon_c=0.25$ and often exceeded $0.30$. These thresholds are task-dependent experimental settings and should not be interpreted as universal safety limits. The reported thresholds were selected as empirical operating points from the present dataset and are intended only to demonstrate separability between controlled and deliberately irregular motion in the tested tasks.


For comparison with the earlier PSM formulation in \cite{tafrishi2022psm}, we also retain a frequency-domain indicator over a sliding window of $N_w$ samples,
\begin{equation}
E_q(k)=
\frac{1}{|\mathcal{B}|}
\sum_{n\in\mathcal{B}}
\left|
\mathcal{F}_{N_w}\!\left\{e_q(k-N_w+1:k)\right\}[n]
\right|,
 q\in\{\theta,\omega\},
\label{eq:spectral_error}
\end{equation}
where $\mathcal{F}_{N_w}\{\cdot\}$ denotes the discrete Fourier transform over the current window and $\mathcal{B}$ is the selected frequency-bin set. This spectral quantity is useful for offline comparison because irregular motion tends to produce broader and larger spectral content than controlled motion. However, by itself it is less convenient for real-time interpretation than the time-domain impedance-inspired index in (\ref{eq:impedance_metric}).


\subsection{Real-Time Evaluation Procedure}

\begin{algorithm}[t!]
\caption{Real-time wrist safety evaluation}
\label{alg:psm_wrist}
\begin{algorithmic}[1]
\Procedure{EvaluateSafety}{$\mathbf{R}_m(k),\dot{\bm{\theta}}_m(k),P_{ij}$}
\State Compute $\theta_g(k)$ from (\ref{eq:theta_g}) and $\omega(k)$ from (\ref{eq:omega_def})
\State Locate the current reduced-state bin $(i,j)$ in the probability map
\State Search a local neighbourhood around $(i,j)$ and select the highest-probability safe-reference state
\State Propagate the wrist PSM in (\ref{eq:psm_wrist}) to obtain $(\bm{\theta}(k),\dot{\bm{\theta}}(k))$
\State Compute $e_{\theta}(k)$ and $e_{\omega}(k)$ from (\ref{eq:time_errors})
\State Compute $e_{\mathrm{imp}}(k)$ from (\ref{eq:impedance_metric})
\State Optionally compute $E_{\theta}(k)$ and $E_{\omega}(k)$ from (\ref{eq:spectral_error}) for offline comparison
\State Classify safety level from (\ref{eq:safety_levels})
\State \Return $e_{\mathrm{imp}}(k),\;S(k)$
\EndProcedure
\end{algorithmic}
\end{algorithm}

\section{Experimental Methodology}
\label{sec:exp_method}

\begin{figure}[t!]
    \centering
    \figinc{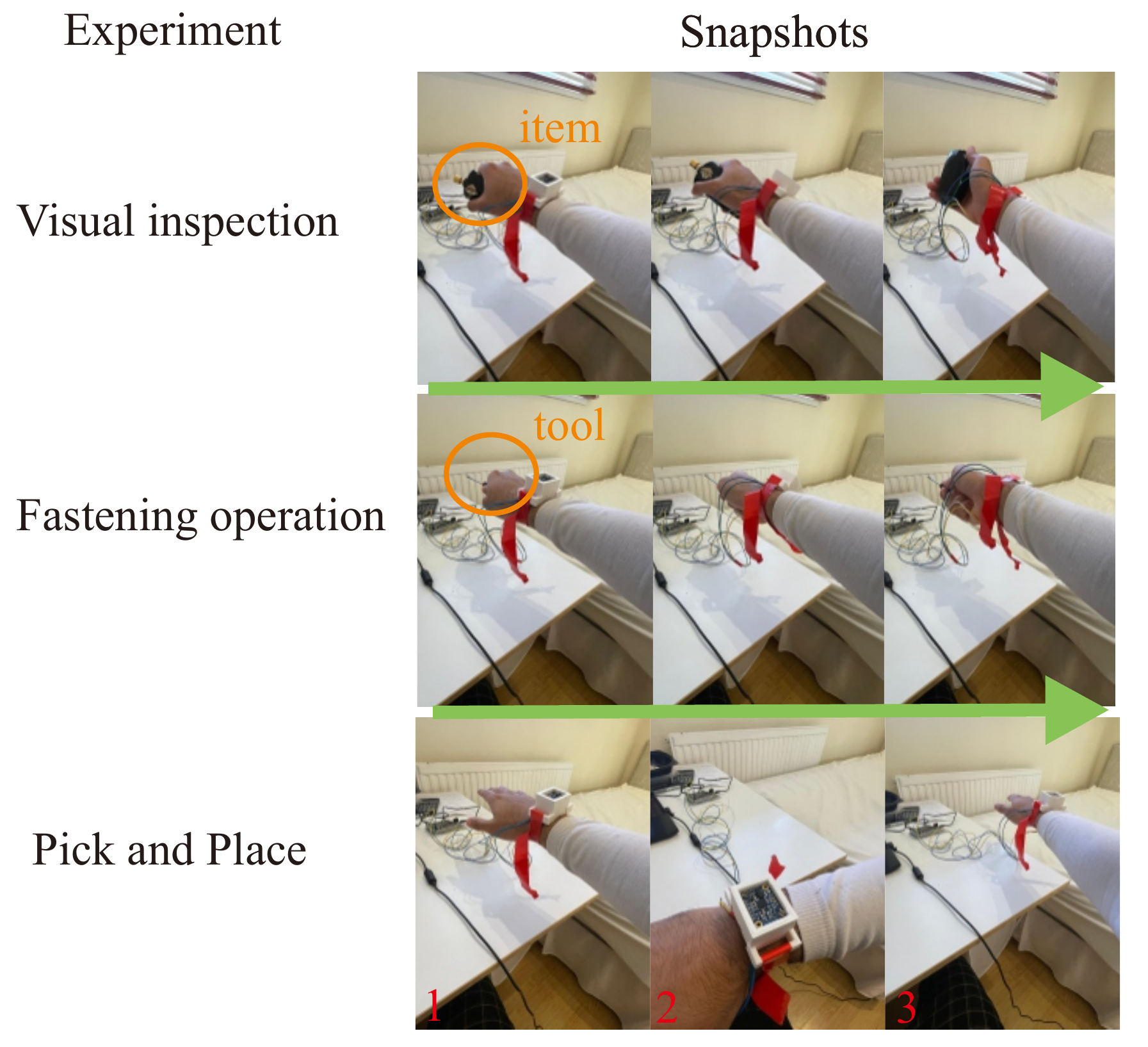}{0.90\linewidth}
    \caption{Experimental tasks used for validation: fastening with a hand tool, visual inspection, and pick-and-place.}
    \label{fig:experiment_snapshots}
\end{figure}

A proof-of-concept validation was conducted in two stages: first, representative single-trial results were used to illustrate the wrist-level behaviour of the proposed PSM; second, the evaluation was extended to a five-participant dataset to test whether the same controlled-versus-irregular separation remained observable across subjects. Three representative manual operations were considered, as shown in Fig.~\ref{fig:experiment_snapshots}:
\begin{enumerate}
    \item \textit{fastening using a hand tool}, involving approach, alignment, wrist rotation, and return,
    \item \textit{visual inspection}, involving repeated scanning of an object with controlled wrist orientation,
    \item \textit{pick-and-place}, involving reach, transfer, and release between two marked locations.
\end{enumerate}

Each task was repeated six times, giving a total of 18 trials. For each task, two trials were performed in a controlled manner, two in a nominal working manner, and two with deliberately irregular execution. The last category included abrupt reversals, overshoot, poor alignment, inconsistent timing, and unnecessary speed bursts in order to generate motion patterns that were qualitatively closer to unsafe or undesirable operation. These labels were introduced to generate diverse motion patterns; therefore, safety was not inferred from speed alone.

The same wrist-level PSM parameters were used in all tasks. The purpose of the experiments was to assess whether the proposed safety indicators could consistently separate controlled task execution from irregular wrist motion across different manual operations. The effective model parameters used throughout the experiments are summarized in Table~\ref{tab:model_params}.

\begin{table}[t!]
\centering
\caption{Effective wrist-level PSM parameters used in all experiments}
\label{tab:model_params}
\begin{tabular}{l c}
\toprule
Parameter & Value \\
\midrule
Effective arm mass $m_a$ & $2~\mathrm{kg}$ \\
Effective pendulum length $l_a$ & $0.2~\mathrm{m}$ \\
Effective base radius $r_a$ & $0.25~\mathrm{m}$ \\
Nominal stiffness $\mathbf{K}_0$ & $[500,\,500,\,1200]^\top$ \\
Nominal damping $\mathbf{B}_0$ & $[40,\,40,\,60]^\top$ \\
Safe threshold $\varepsilon_s$ & $0.15$ \\
Caution/unsafe-like threshold $\varepsilon_c$ & $0.25$ \\
Default online weighting & $P^*=0.4,\;D^*=0.6$ \\
\bottomrule
\end{tabular}
\end{table}

\begin{figure}[t!]
    \centering
    \figinc{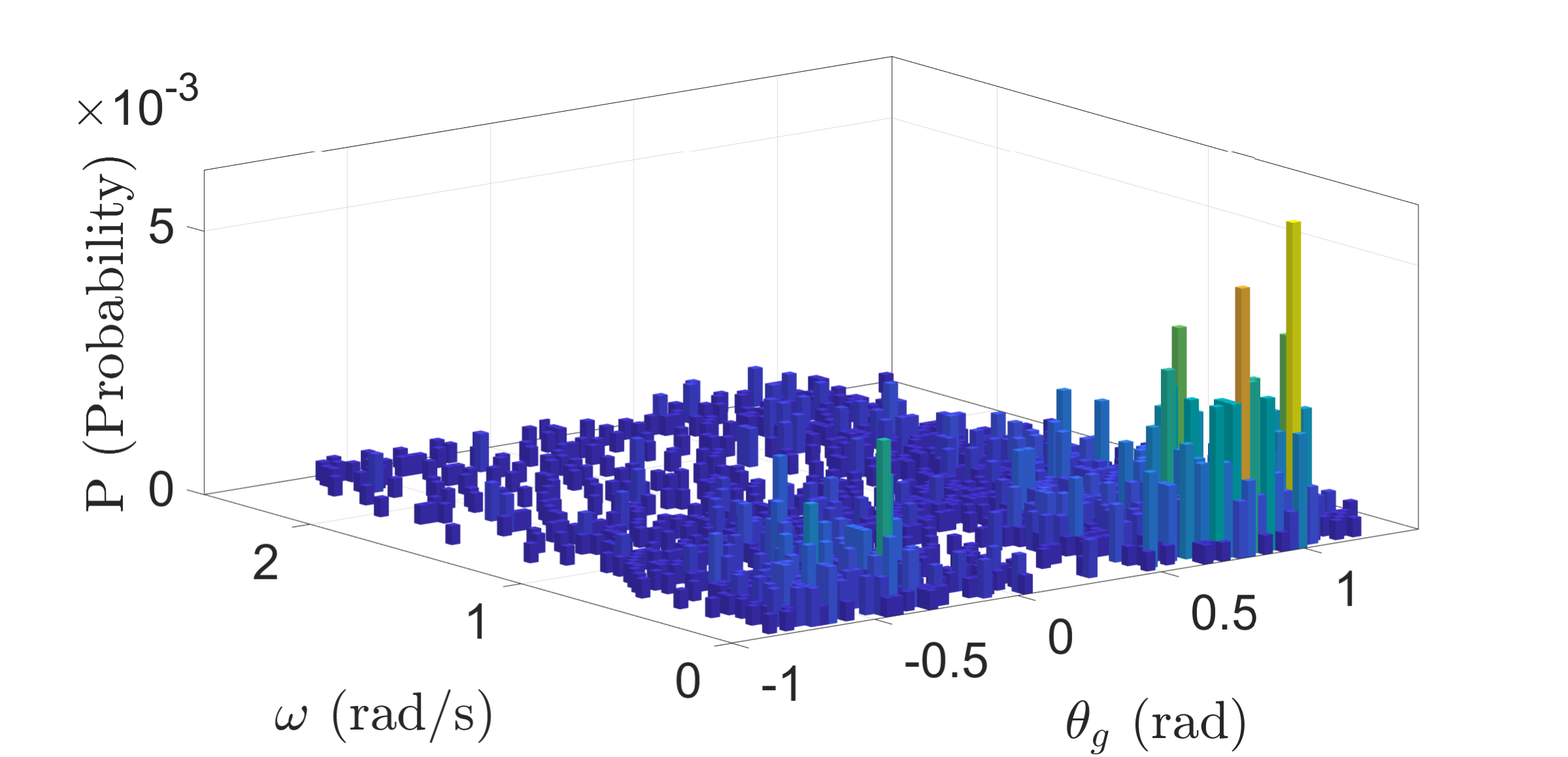}{0.92\linewidth}
    \caption{Example probability distribution over the reduced wrist-motion state $(\theta_g,\omega)$ for one of the manufacturing tasks. High-density regions represent common and repeatable wrist states, while low-density regions correspond to uncommon or irregular motion.}
    \label{fig:Probability_plot}
\end{figure}

A task-specific probability map was constructed from the reduced variables $(\theta_g,\omega)$ of all six trials for each task. Figure~\ref{fig:Probability_plot} shows an example of the resulting probability distribution. Peaks in the distribution correspond to common and repeatable wrist states, whereas sparse regions indicate less regular motion. In the online evaluation, the probability map is not used as a binary classifier but as a compact reference structure from which the most likely safe continuation of the motion is inferred.

\section{Results and Discussion}
\label{sec:motion_study}

\begin{figure}[t!]
    \centering
    \figinc{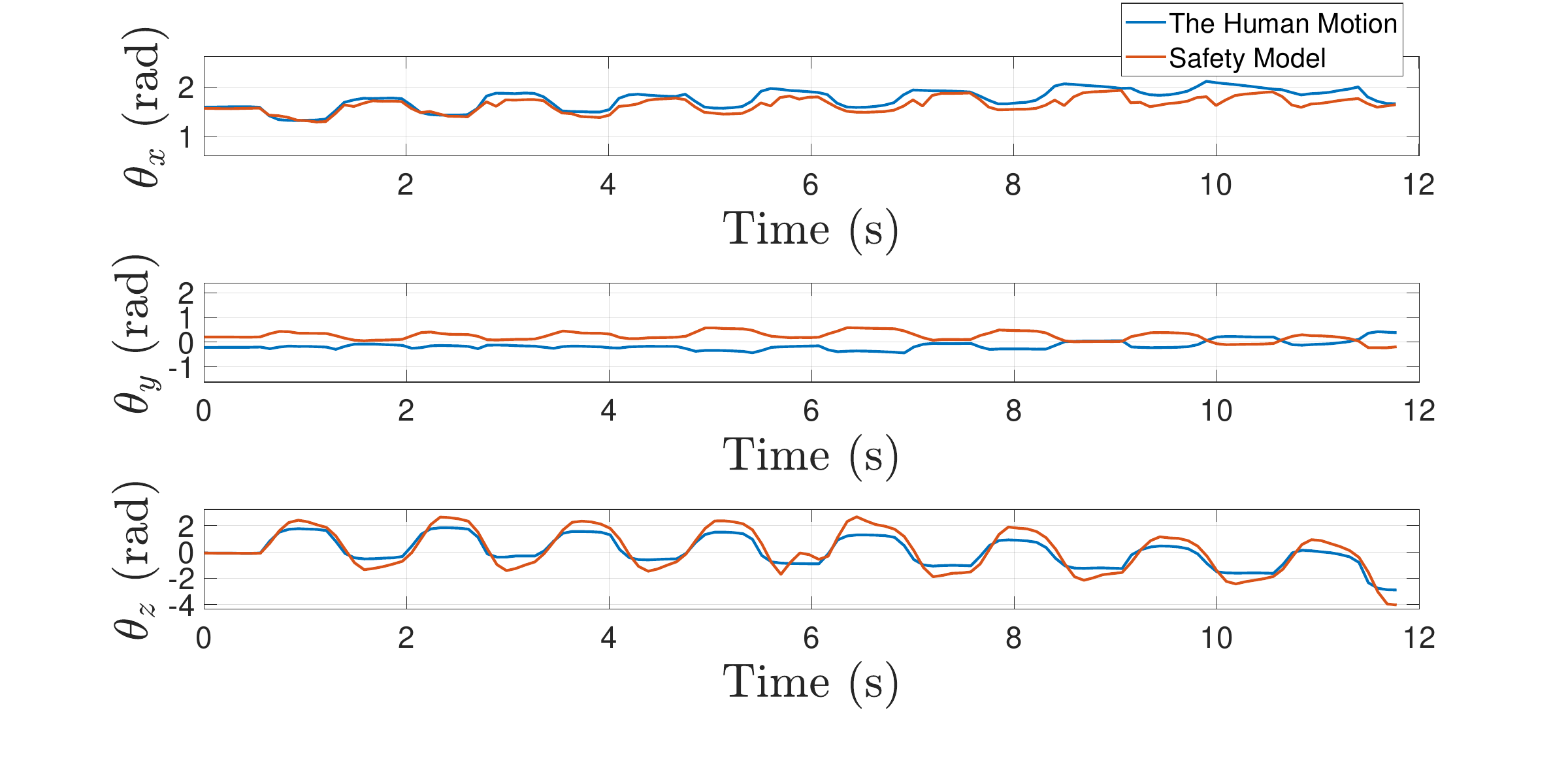}{1\linewidth}\hfill
    \figinc{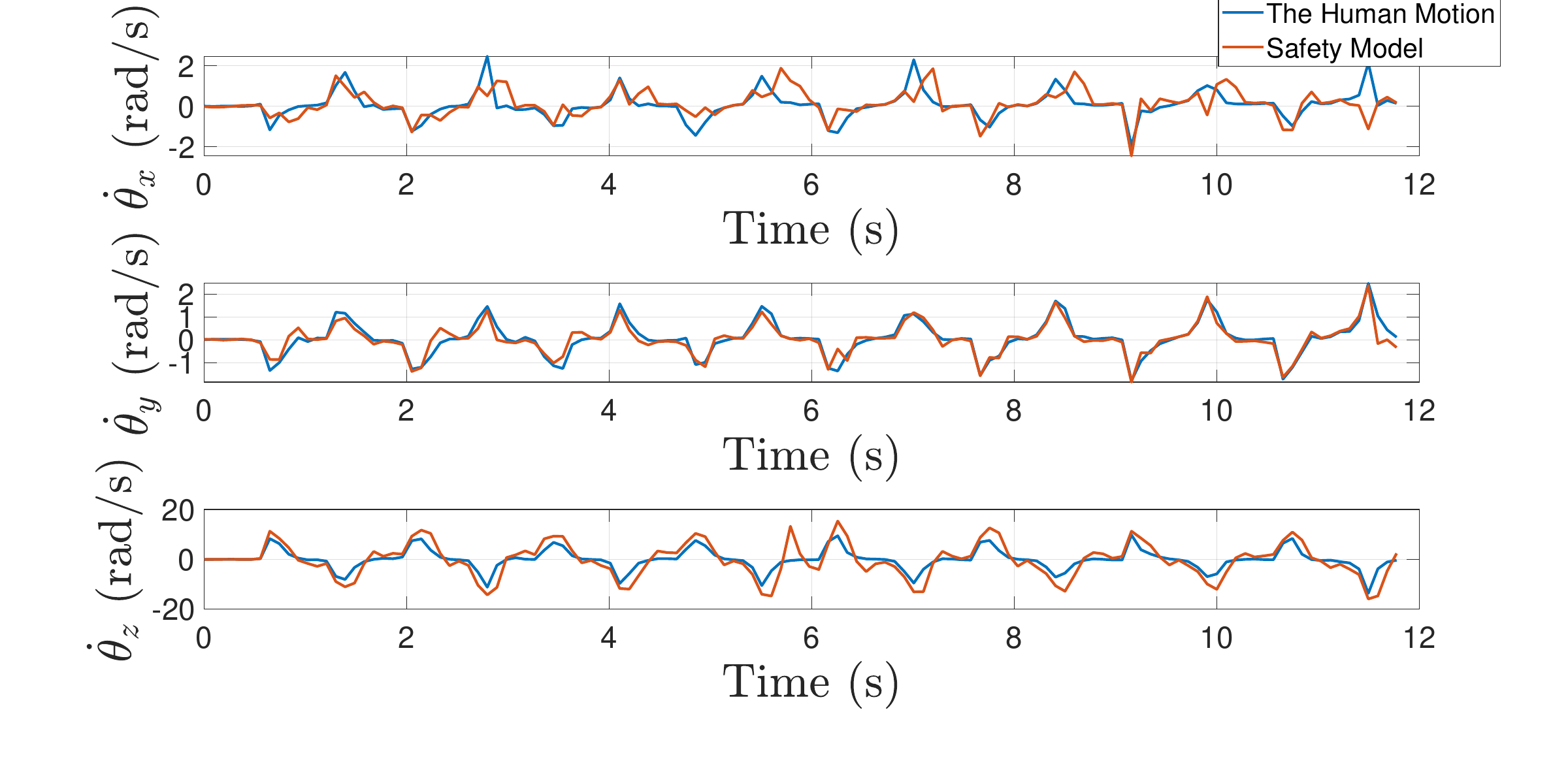}{1\linewidth}
    \caption{Fastening task under controlled execution. The measured wrist motion remains close to the task-consistent PSM prediction, leading to comparatively small tracking discrepancies.}
    \label{fig:fastening_normal}
\end{figure}

\begin{figure}[t!]
    \centering
    \figinc{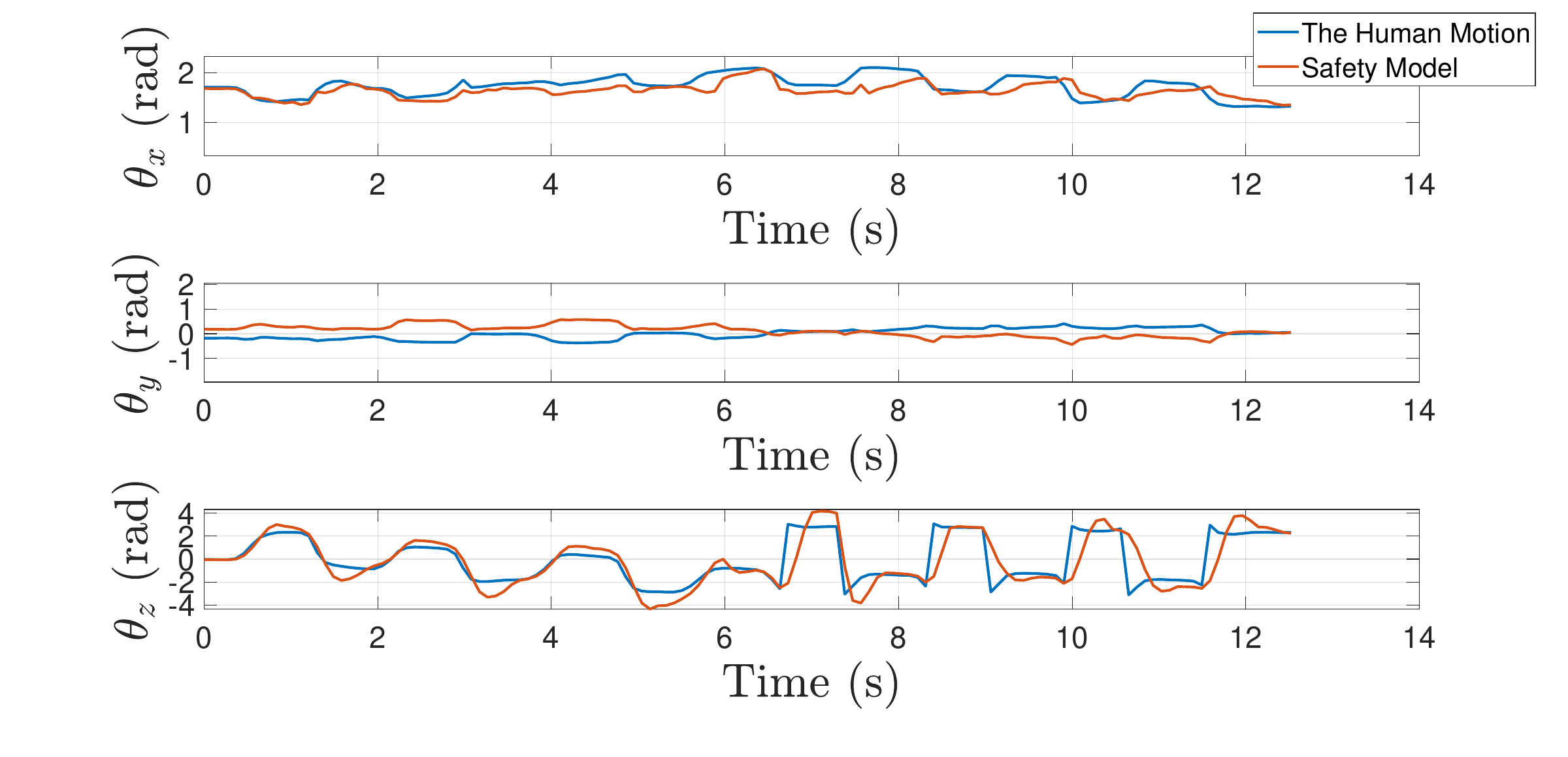}{1\linewidth}\hfill
    \figinc{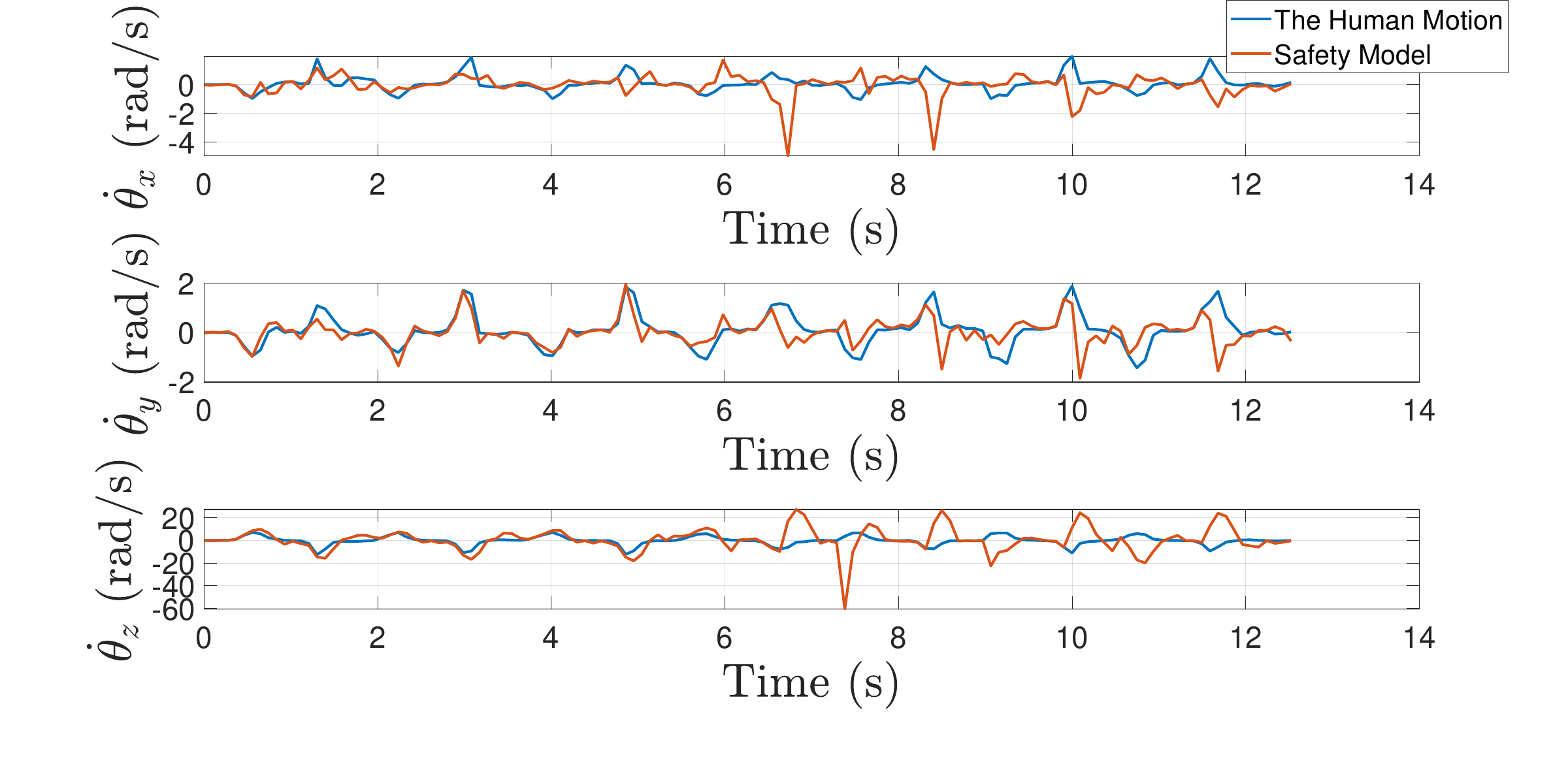}{1\linewidth}
    \caption{Fastening task under deliberately irregular execution. Abrupt reversals, oscillatory corrections, and inconsistent wrist timing produce larger deviations from the predicted safe motion.}
    \label{fig:fastening_unsafe}
\end{figure}

The proposed wrist-level PSM was first examined qualitatively through representative single-trial task executions and then through an aggregated multi-participant evaluation. The fastening task gives the clearest qualitative example because it requires repeated wrist rotation with relatively tight orientation control. In the controlled trial of Fig.~\ref{fig:fastening_normal}, the measured wrist motion remains close to the PSM-predicted safe trajectory, with smooth rotational evolution and limited corrective action. In contrast, the deliberately irregular trial in Fig.~\ref{fig:fastening_unsafe} exhibits abrupt reversals, oscillatory corrections, and stronger axis-to-axis variability, which directly increase the deviation between the measured wrist state and the predicted safe state. Accordingly, the labels \emph{unsafe-like} or \emph{irregular} in this paper refer to deliberately non-compliant wrist motion patterns used for proof-of-concept evaluation, rather than independently validated safety incidents. The same qualitative trend was also observed in visual inspection and pick-and-place, although the absolute motion envelope is naturally broader in pick-and-place because the hand repeatedly transitions between reach, transfer, and release phases.

\begin{figure}[t!]
    \centering
    a)\,\figinc{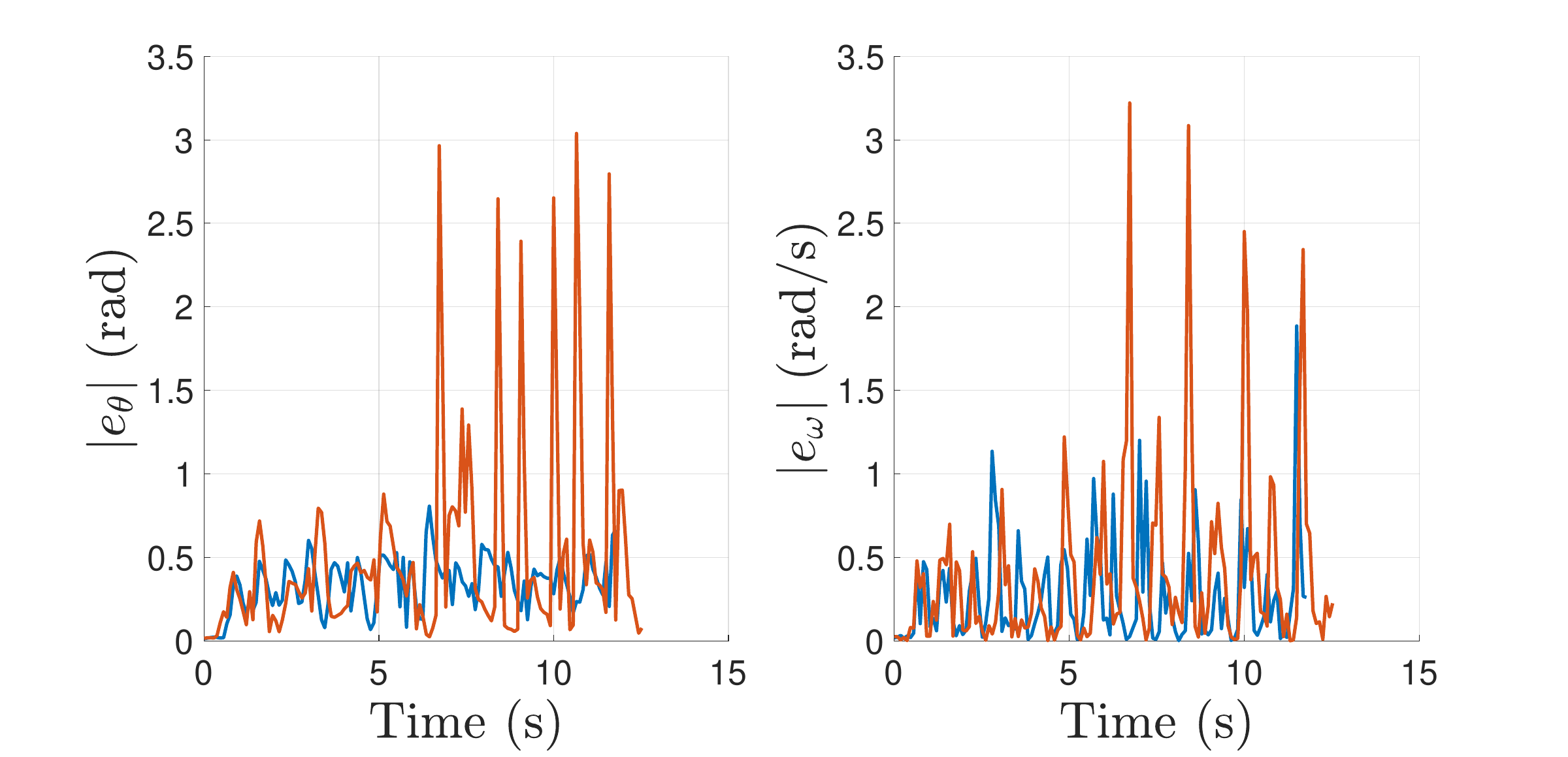}{0.46\linewidth}\hfill
    \figinc{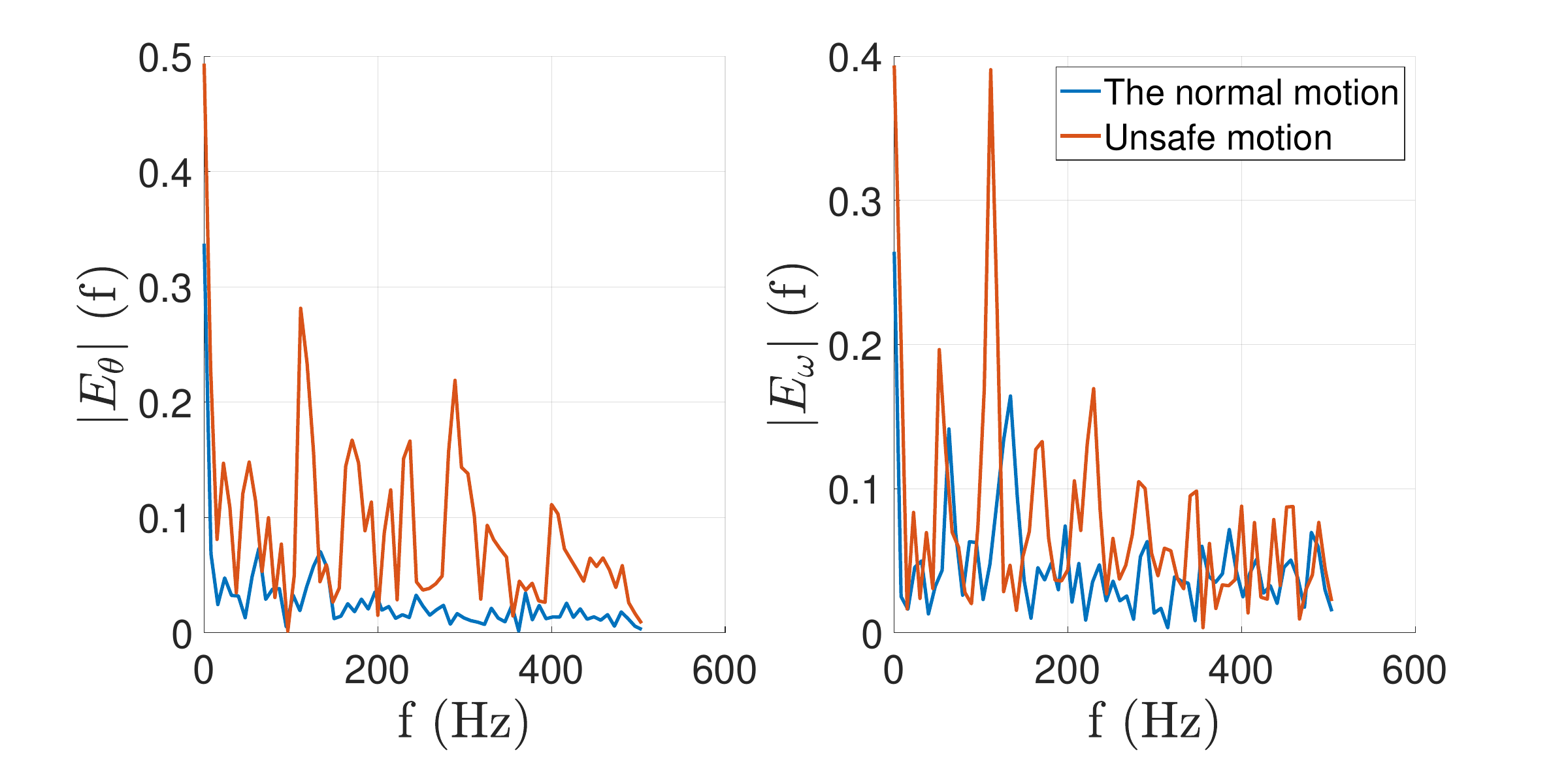}{0.46\linewidth}\\[1mm]
    b)\,\figinc{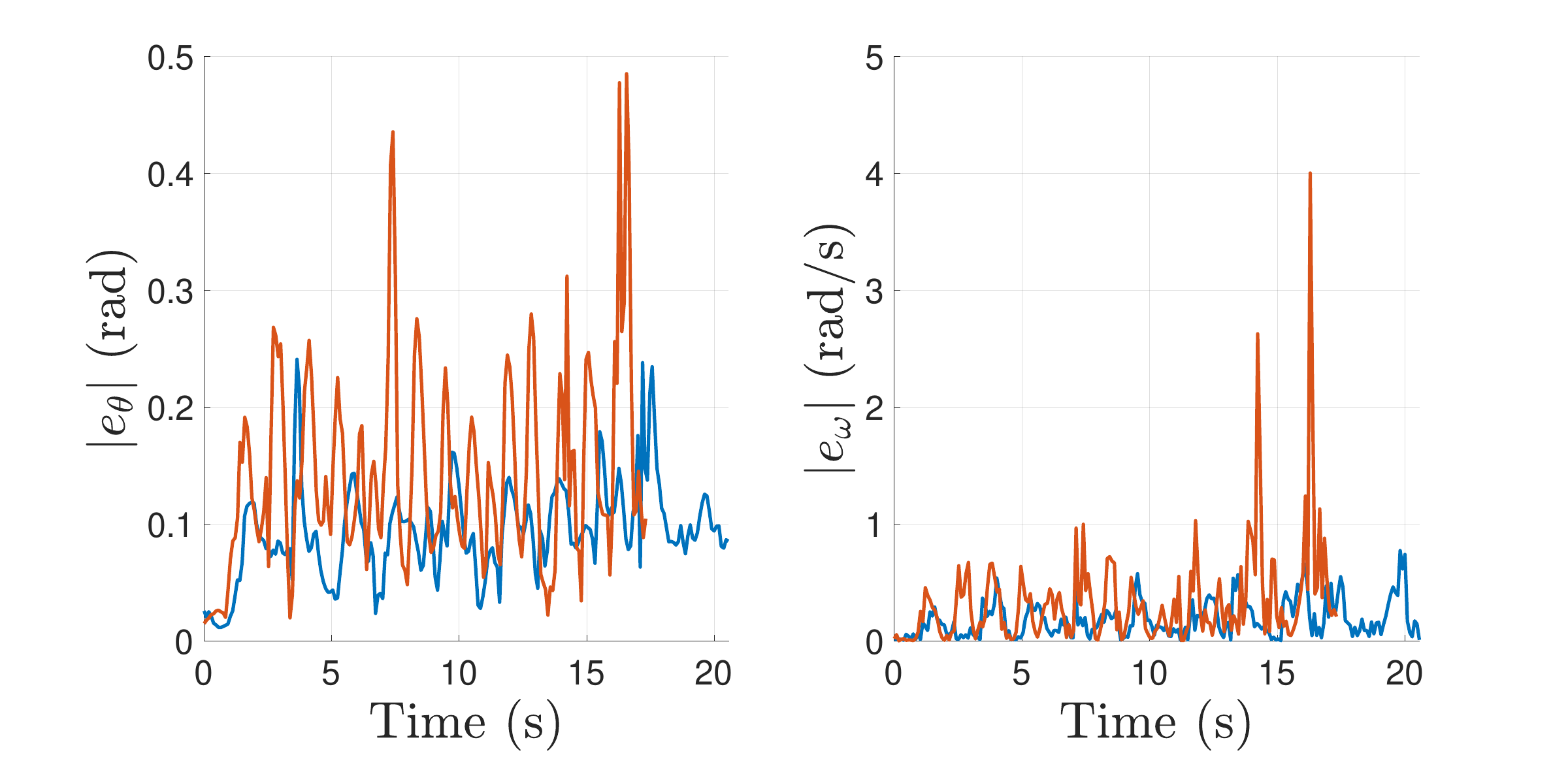}{0.46\linewidth}\hfill
    \figinc{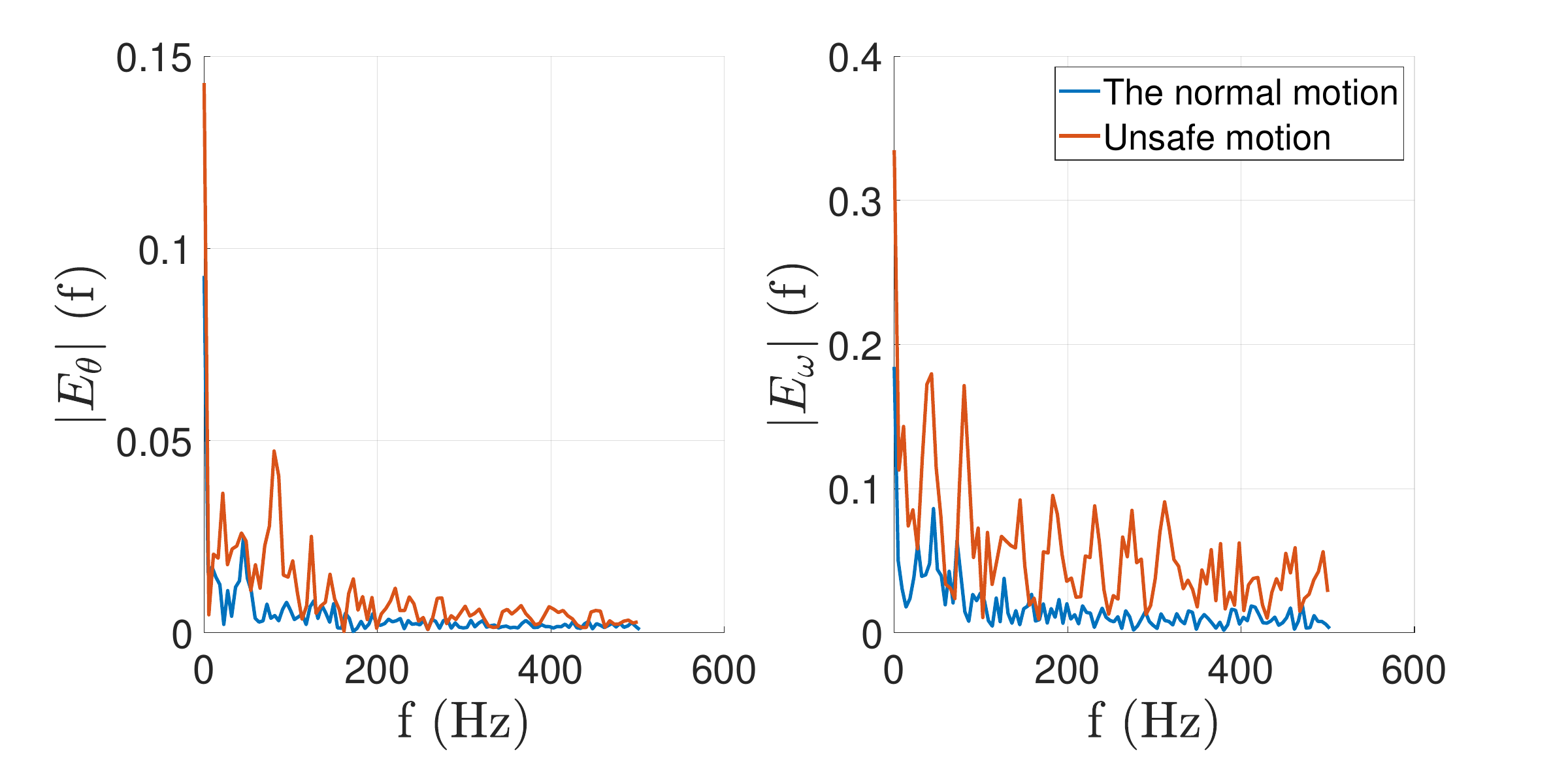}{0.46\linewidth}\\[1mm]
    c)\,\figinc{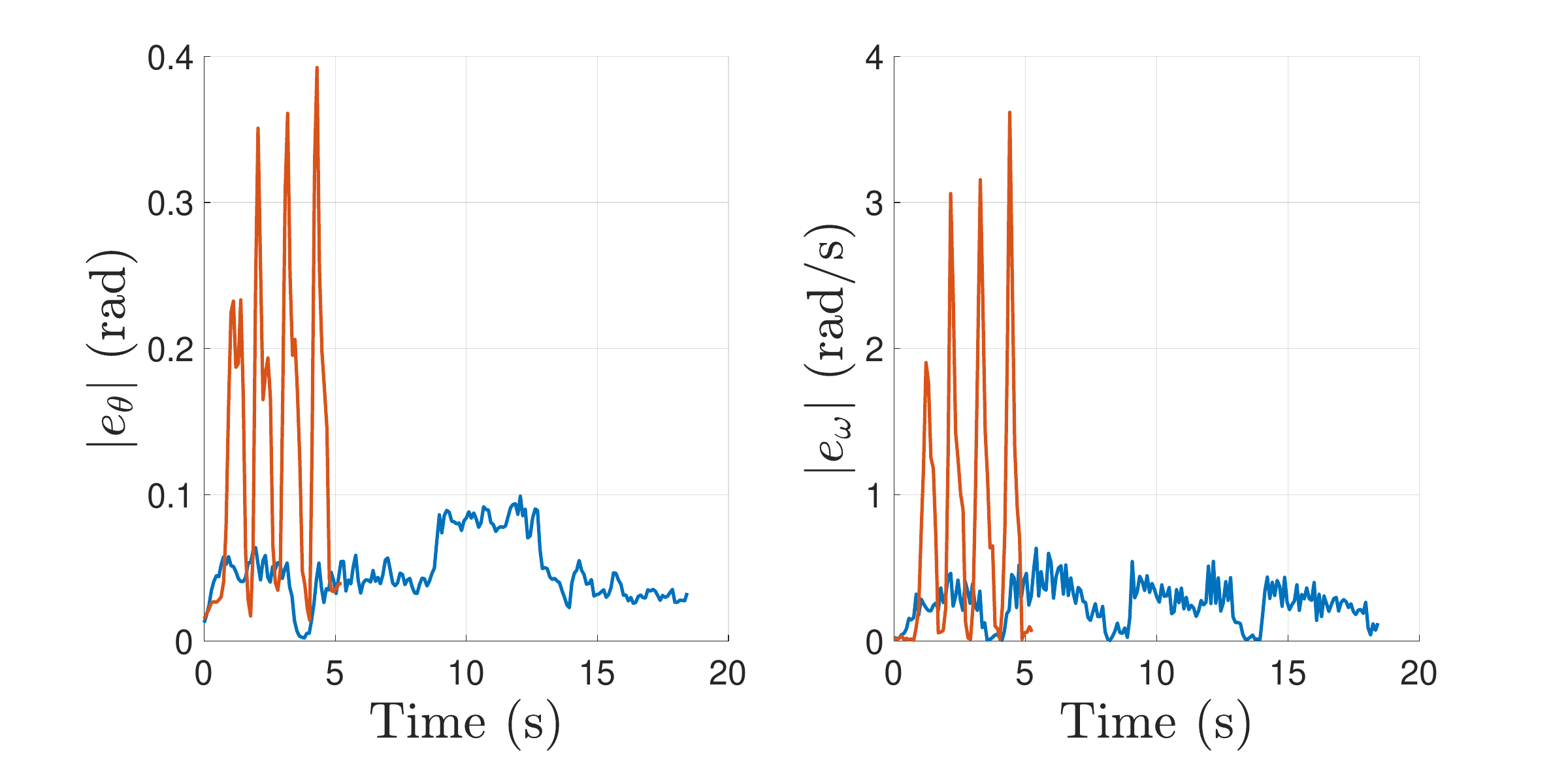}{0.46\linewidth}\hfill
    \figinc{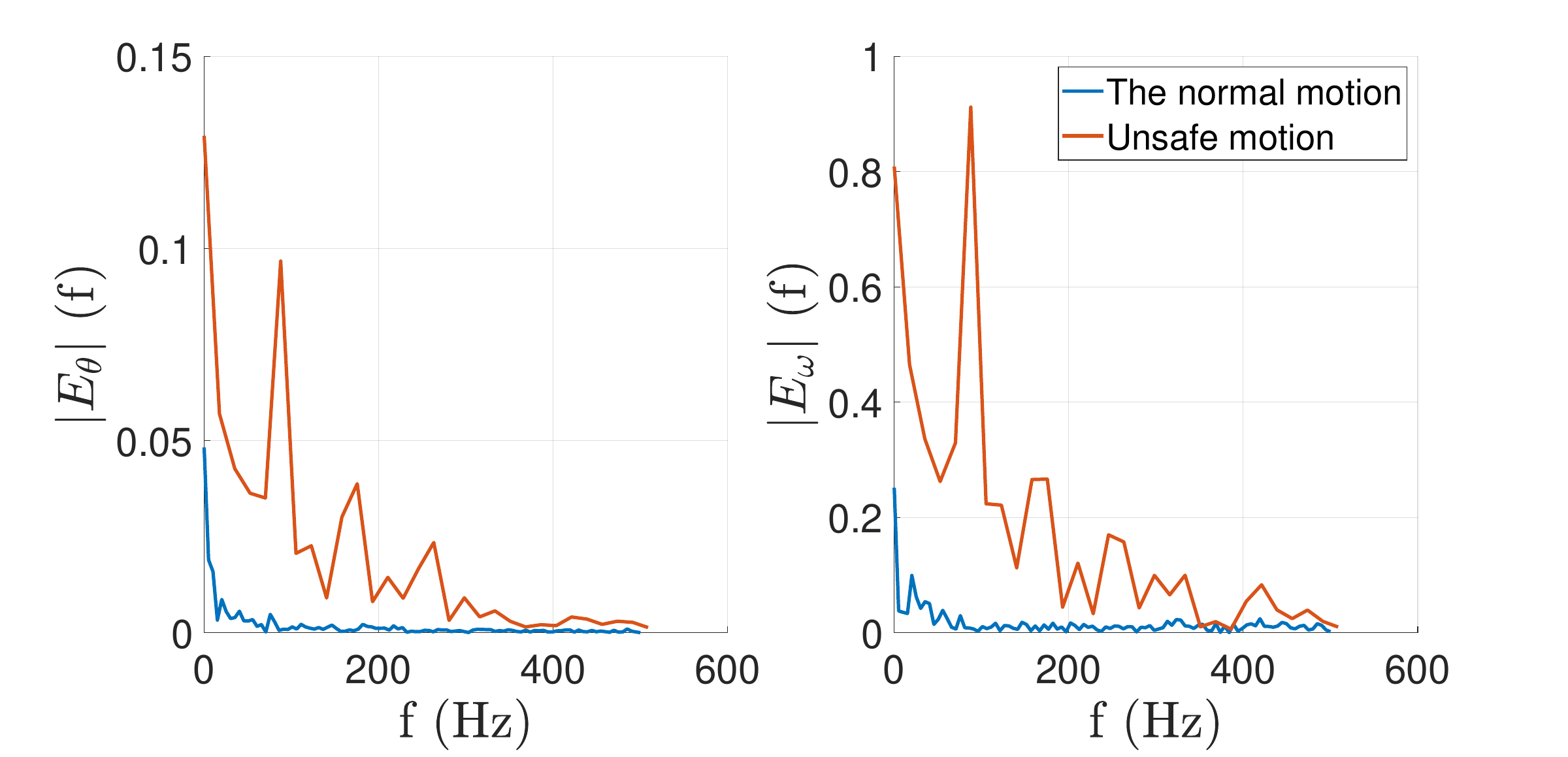}{0.46\linewidth}
    \caption{RMS error traces and corresponding spectral comparisons for (a) fastening, (b) visual inspection, and (c) pick-and-place. In all three tasks, deliberately irregular motions generate larger deviations than controlled task execution.}
    \label{fig:rms_overall}
\end{figure}

\begin{figure}[t!]
    \centering
    \figinc{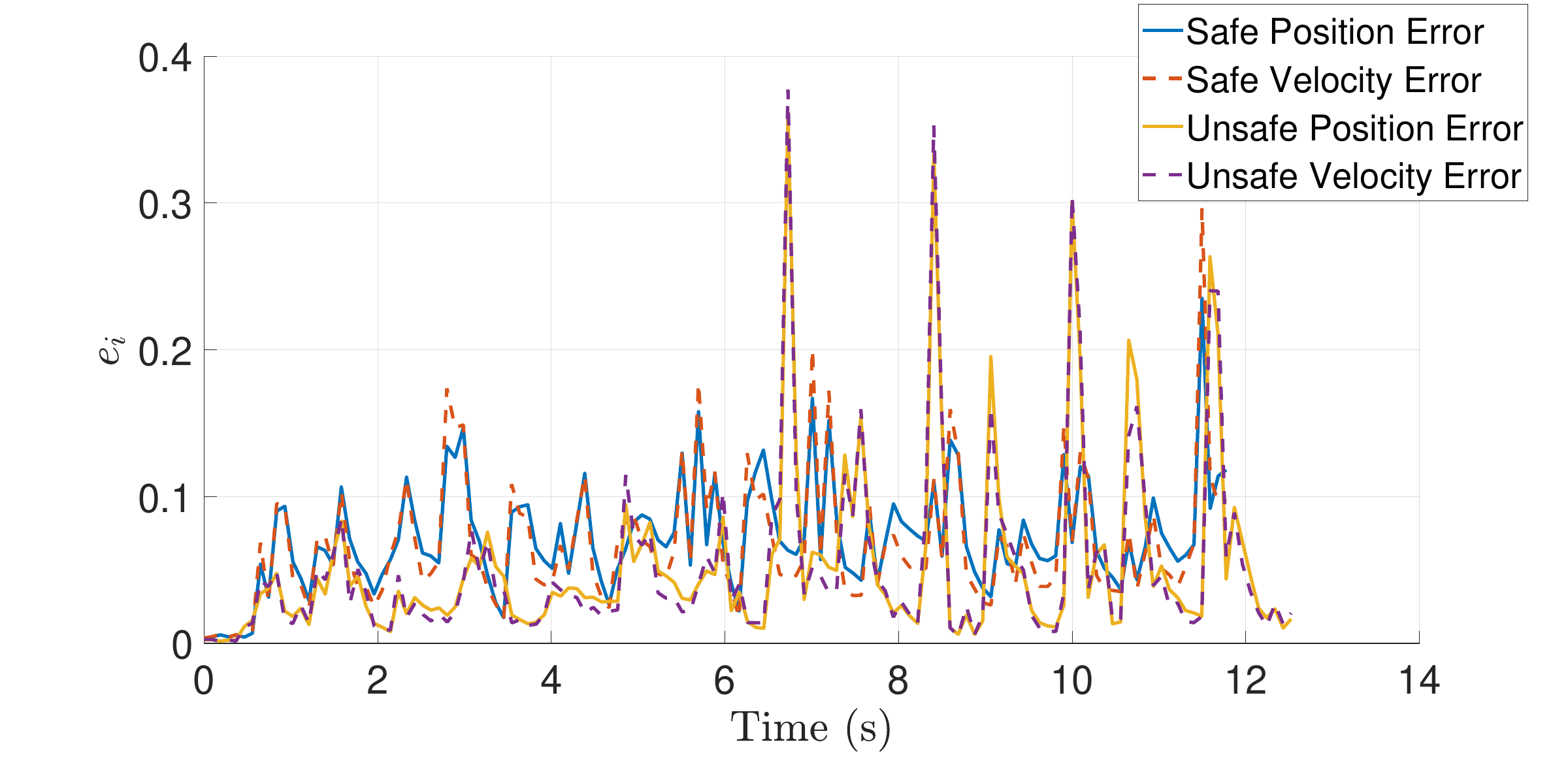}{0.5\linewidth}\hfill
    \figinc{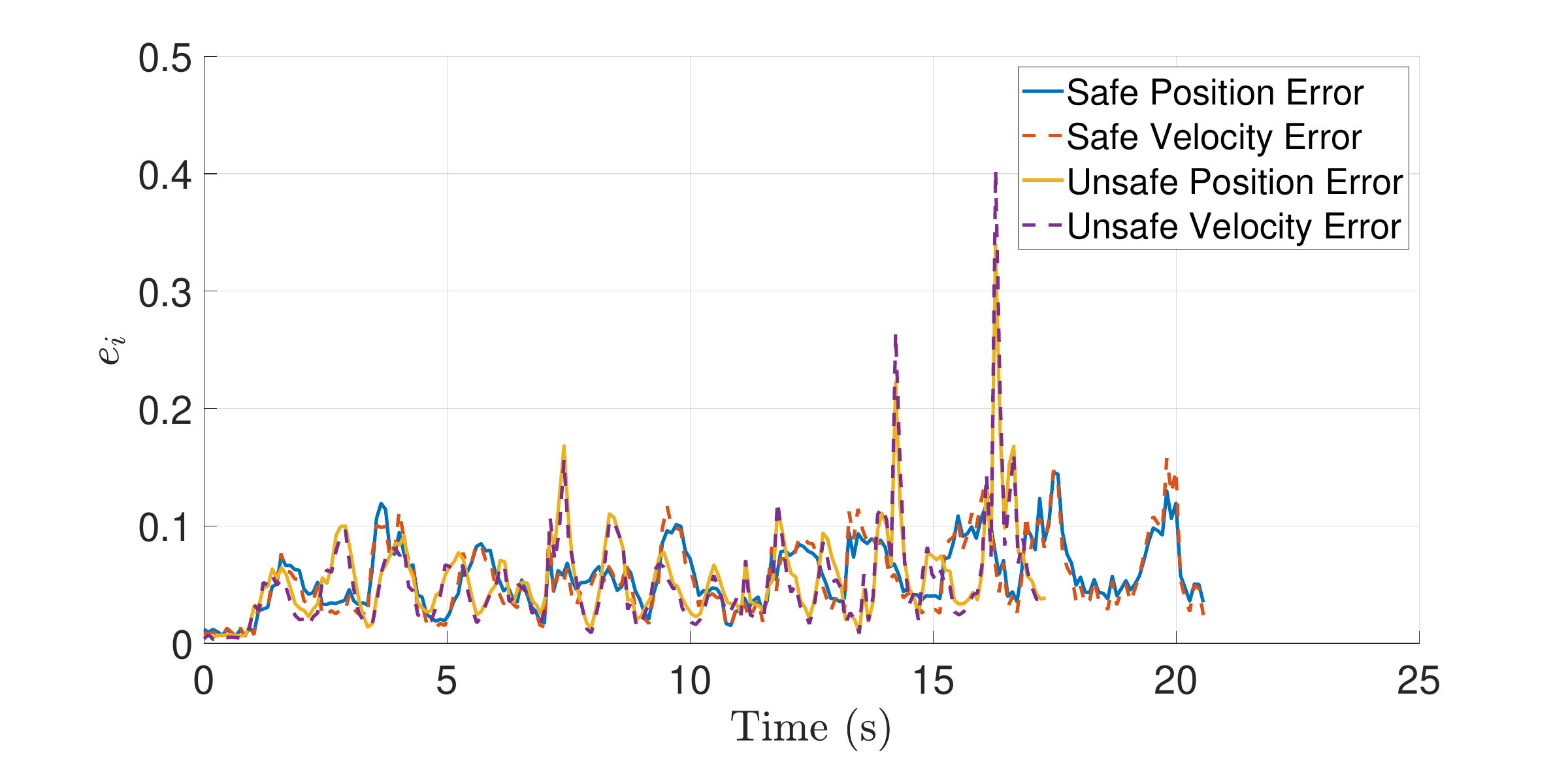}{0.5\linewidth}\\
    (a)\hspace{4.1cm}(b)\\[1mm]
    \figinc{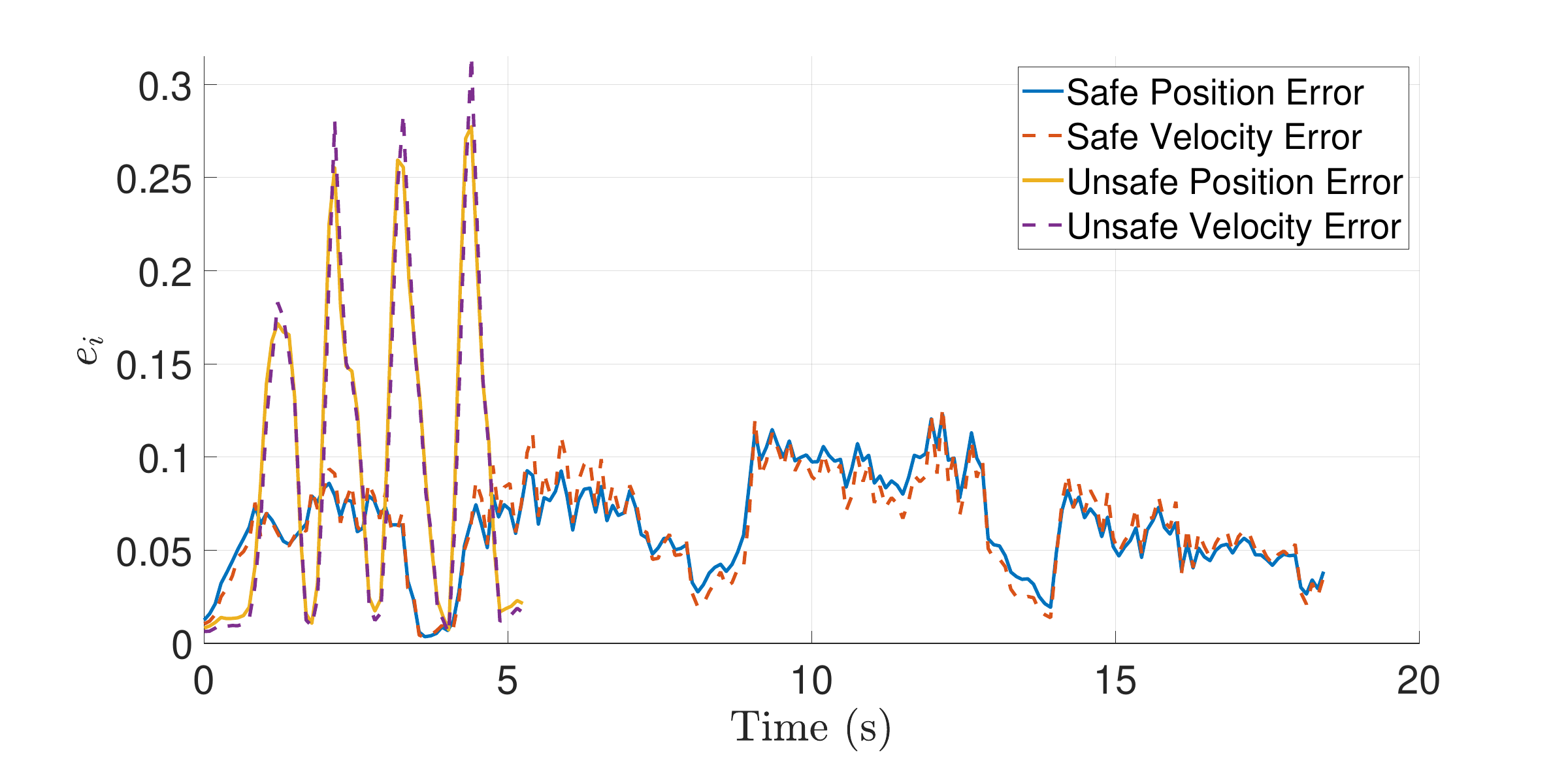}{0.5\linewidth}\\
    (c)
    \caption{Risk assessment using the impedance-inspired error metric for (a) fastening, (b) visual inspection, and (c) pick-and-place. The two curves correspond to position-priority and velocity-priority weighting.}
    \label{fig:impedance_tasks}
\end{figure}

This behaviour is also reflected in the error-domain comparisons. Figure~\ref{fig:rms_overall} shows that deliberately irregular executions produce larger RMS deviations than controlled executions for both angular orientation and angular velocity across fastening, visual inspection, and pick-and-place. The associated spectral plots reveal the same overall trend, confirming that irregular behaviour is not only larger in magnitude but also richer in temporal fluctuation. However, the frequency-domain representation is more suitable as an offline comparative baseline than as the main online safety variable because it requires a windowed spectral computation and does not immediately indicate the exact time instant at which the motion becomes unsafe-like. For this reason, the impedance-inspired metric in Fig.~\ref{fig:impedance_tasks} is retained as the primary online indicator. In the present data, controlled executions generally remain in a relatively low normalized range, whereas deliberately irregular executions generate clear peaks during abrupt corrections and poorly coordinated wrist actions.

\begin{figure}[t!]
    \centering
    \figinc{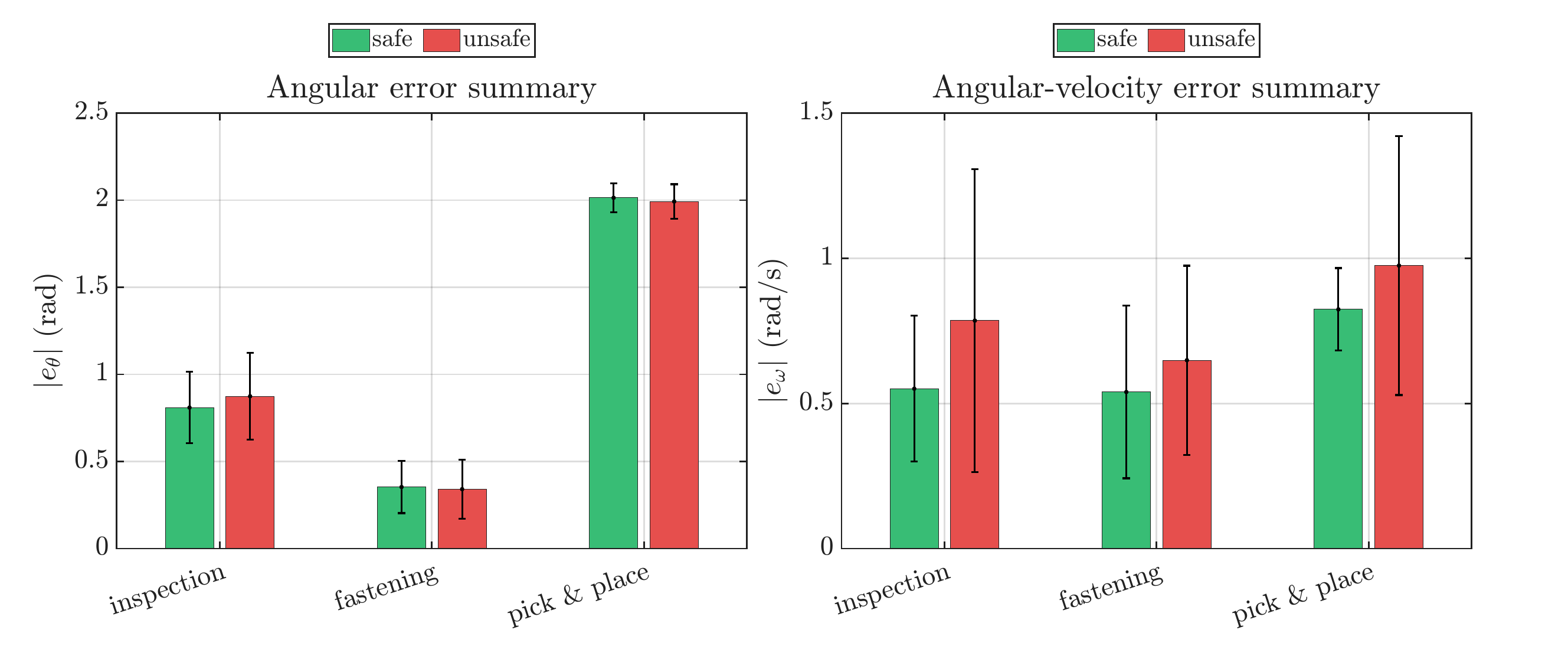}{0.98\linewidth}\\[1mm]
    \figinc{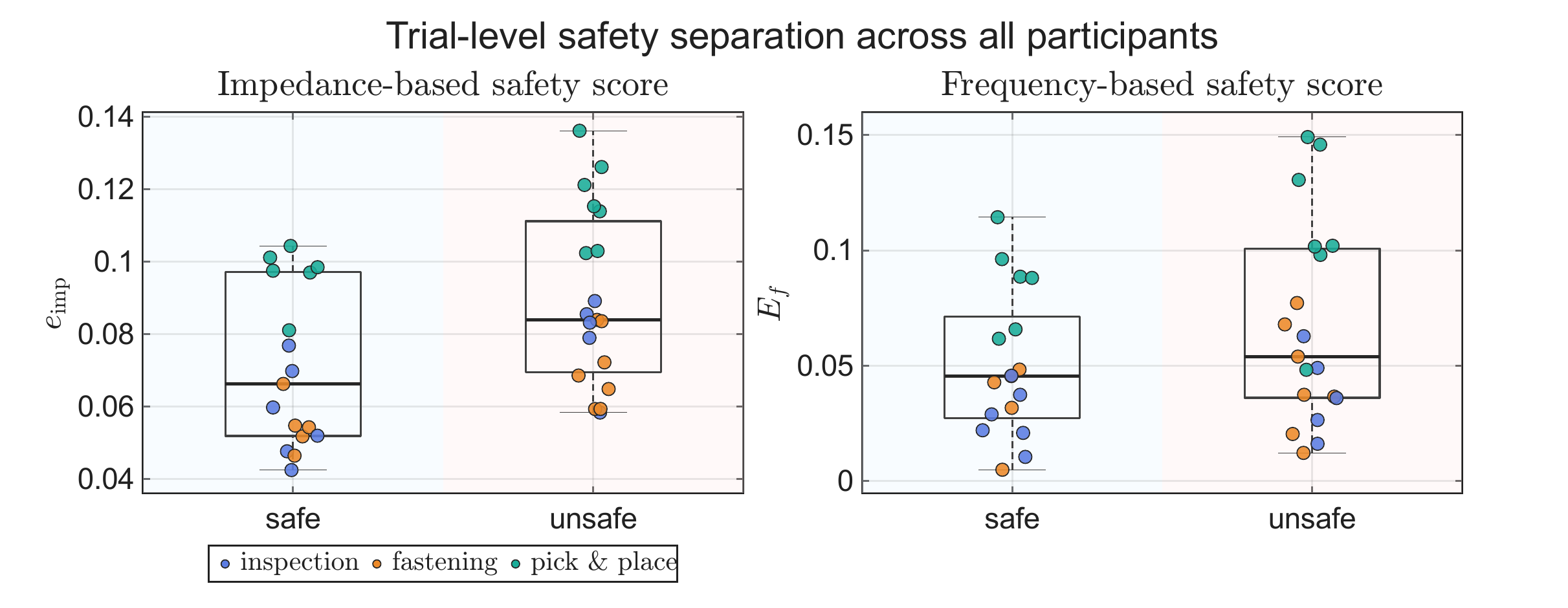}{0.98\linewidth}\\[1mm]
    \figinc{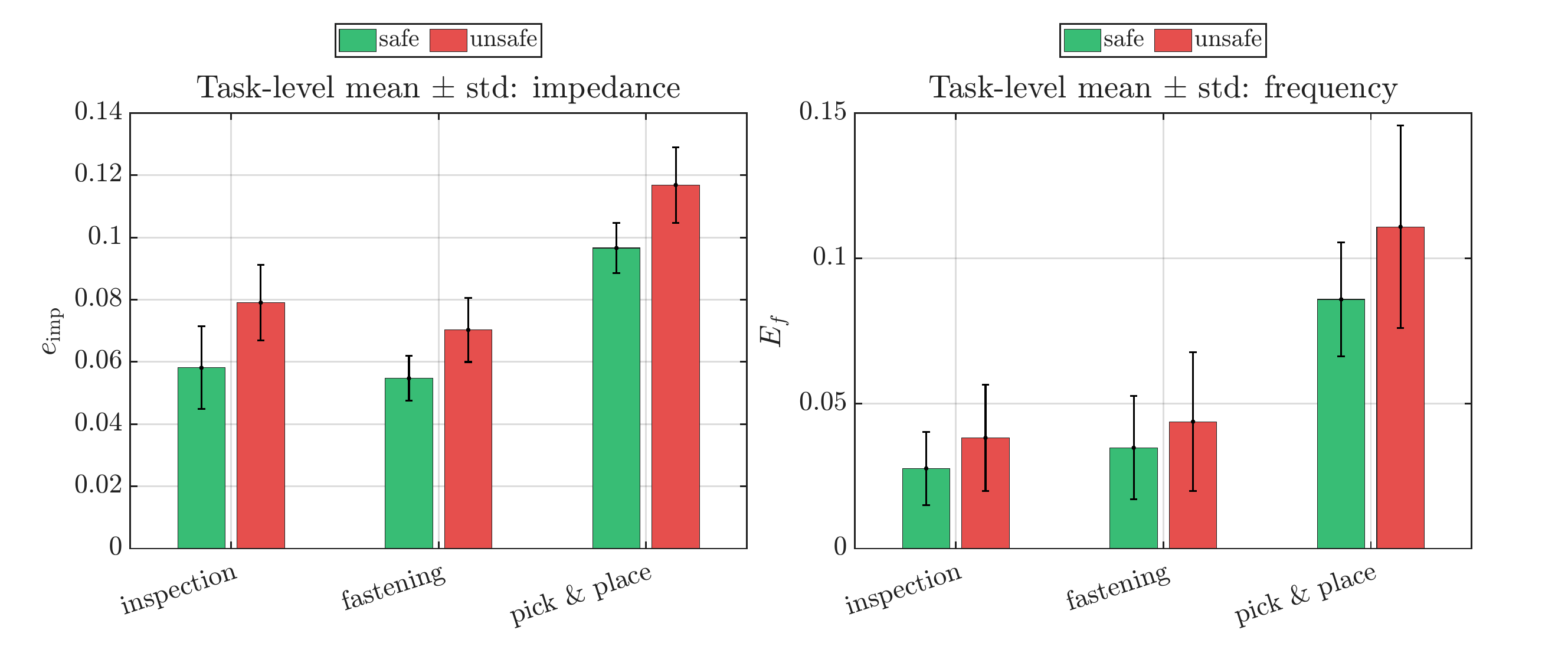}{0.98\linewidth}
    \caption{Multi-participant evaluation of the proposed wrist-level safety monitor: top, trial-level separation of safe and unsafe motions for the impedance- and frequency-based scores; middle, task-wise mean $\pm$ std of the two safety scores; bottom, task-wise mean $\pm$ std of the angular and angular-velocity errors.}
    \label{fig:multi_participant_stats}
\end{figure}

To verify that the separation observed in the illustrative trials is not an isolated case, the evaluation was extended to a multi-participant dataset comprising five healthy participants aged 18--26 years and 30 recordings spanning controlled and deliberately irregular executions of inspection, fastening, and pick-and-place tasks. All participants were tested under the same acquisition protocol, including a common starting posture and repeated trial structure, which reduced procedure-dependent variation and improved comparability across recordings. At the aggregate level, controlled trials remained concentrated in the lower motion-intensity range, whereas irregular trials showed a consistent upward shift in both the impedance-based and frequency-based indicators. Collectively, these results confirm that the proposed framework captures a reproducible distinction between safe and irregular wrist execution across participants and therefore supports the multi-participant comparisons reported in Fig.~\ref{fig:multi_participant_stats}.

The top and middle parts of Fig.~\ref{fig:multi_participant_stats} show that both safety scores preserve a consistent upward shift from the safe class to the unsafe class across tasks. At trial level, the impedance score shows a clearer separation in central tendency, with the unsafe box shifted upward relative to the safe box, while the frequency score exhibits a broader spread and a stronger upper tail for unsafe motions. At task level, the impedance score rises from approximately $0.058$ to $0.079$ in inspection, from about $0.055$ to $0.070$ in fastening, and from about $0.096$ to $0.116$ in pick-and-place. The corresponding frequency score increases from roughly $0.027$ to $0.037$ in inspection, from about $0.033$ to $0.043$ in fastening, and from about $0.085$ to $0.111$ in pick-and-place. These results indicate that pick-and-place produces the highest absolute safety scores in both methods, which is consistent with its broader motion envelope, whereas inspection and fastening show smaller absolute values but still maintain clear safe--unsafe separation. The bottom part of Fig.~\ref{fig:multi_participant_stats} provides the physical interpretation behind that separation. The mean angular error increases only modestly from safe to unsafe in inspection, from about $1.05$ to $1.16$ rad, and in fastening, from about $0.42$ to $0.46$ rad, while pick-and-place remains nearly unchanged at roughly $0.70$--$0.71$ rad. In contrast, the angular-velocity error gives a much clearer class separation: it rises from about $0.28$ to $0.45$ rad/s in inspection, from about $0.42$ to $0.66$ rad/s in fastening, and from about $0.25$ to $0.33$ rad/s in pick-and-place. This behaviour supports the use of velocity-priority weighting in the impedance-inspired score: for a wrist-mounted IMU, hazardous-like behaviour is manifested more consistently through temporal irregularity and abrupt corrective action than through a large absolute orientation offset alone.

Overall, the revised aggregated results strengthen the same conclusion suggested by the single-trial observations. Both metrics distinguish controlled from deliberately irregular motion, but the impedance-inspired score remains the more practical online indicator because it preserves safe--unsafe separation while remaining directly tied to time-domain motion deviation. The frequency-based score is still useful as a complementary offline validation measure, especially for highlighting the broader spread and stronger high-variability tail of unsafe motions. The present thresholds should therefore be interpreted as empirical operating points for the tested tasks rather than universal safety limits, and broader validation across larger cohorts and other sensing modalities remains part of future work.


\section{Conclusion}
\label{sec:conclusions}

This paper presented a wrist-level adaptation of the Predictive Safety Model for evaluating human arm operations using only a wrist-mounted IMU. The approach combines the spring-damper-mass predictive model with an impedance-inspired safety index for online monitoring, while retaining spectral error analysis as a complementary offline reference.

Experiments on fastening, visual inspection, and pick-and-place tasks showed that the proposed method can distinguish controlled execution from deliberately irregular wrist motion across different task conditions. In the present dataset, the velocity-priority weighting provided the most stable online behaviour, indicating that temporal irregularity in wrist motion is often more informative than absolute orientation deviation alone.

Overall, the work provides a practical extension of the earlier PSM framework toward wrist-based safety evaluation in collaborative manufacturing scenarios. Future work will focus on larger multi-participant validation, task-adaptive threshold selection, and comparison with additional real-time safety monitoring approaches for human-robot collaboration.


\balance
\bibliographystyle{IEEEtran}
\bibliography{references.bib}

\end{document}